\documentclass{article}

\usepackage{arxiv}

\usepackage{hyperref}
\usepackage[T1]{fontenc}
\usepackage{graphicx}
\usepackage{amssymb,amsmath,amsthm}
\usepackage{mathtools}
\usepackage{multirow}
\usepackage{booktabs}	
\usepackage{float}	
\usepackage{amsmath,amsfonts,latexsym,amssymb}
\usepackage[mathscr]{euscript}
\usepackage{mathtools}
\usepackage{subfigure}
\usepackage{cite}
\usepackage{enumitem}
\usepackage{bm}
\usepackage{color,soul} 
\usepackage{array}
\usepackage[ruled,linesnumbered]{algorithm2e}
\usepackage{siunitx} 
\usepackage{cancel} 
\usepackage{comment} 

\usepackage{diagbox}

\newlength{\figwidth}
\newlength{\subfigwidth}
\newlength{\subfigheight}
\newcolumntype{d}[1]{D{.}{.}{#1}}

\title{Simulation free reliability analysis: A physics-informed deep learning based approach}

\author{
 Souvik Chakraborty \\
  Department of Applied Mechanics,\\
  Indian Institute of Technology Delhi,\\
  New Delhi, India.\\
  \texttt{csouvik41@gmail.com} \\
   }
 
\begin{document}
\maketitle
\begin{abstract}
This paper presents a {\it simulation free}  framework for solving reliability analysis
problems.
The method proposed is rooted in a recently 
developed deep learning approach, referred to as the physics-informed neural network.
The primary idea is to learn the neural network parameters directly from the physics of the problem.
With this, the need for running simulation and 
generating data is completely eliminated.
Additionally, the proposed approach also satisfies physical laws such as invariance properties and conservation laws associated with the problem.
The proposed approach is used for solving three benchmark reliability analysis problems. 
Results obtained illustrates that the proposed
approach is highly accurate.
Moreover, the primary bottleneck of solving reliability analysis problems, i.e., running expensive simulations to generate data,
is eliminated with this method.
\end{abstract}


	
\section{Introduction}
\label{sec:intro}
The primary aim of the domain {\it reliability analysis} is to estimate
the probability of failure of a system.
Theoretically, this is straightforward to formulate as it is, in
essence, a multivariate integration problem \cite{haldar2000reliability,haldar2000probability}.
However, from a practical point-of-view, computing probability of 
failure is often  a daunting task.
Often there exists no closed form solution for the 
multivariate integral and one has to rely on numerical integration
techniques. 
Also, the failure domain, over which the multivariate integration is
to be carried out, is often irregular.

Monte Carlo simulation (MCS) \cite{Thakur1978monte,Rubinstein1981simulation} is perhaps the most straightforward method for reliability analysis.
In this method, the multivariate integral is approximated by 
using large number of samples drawn from the probability distribution
of the input variables.
Simulation is carried out corresponding to each of the drawn samples
and one checks whether failure has occurred or not.
Unfortunately, the convergence rate of MCS in number of samples
is very slow. 
Large number of samples are needed to achieve
a converged solution.
Consequently, MCS is computationally cumbersome.
To address this issue, researchers over the years have developed
methods that are improvements over the {\it vanilla MCS} discussed
above. 
Such methods include importance sampling (IS) \cite{Au1999a,li2005curse}, subset simulation (SS) \cite{Au2001estimation,Au2014engineering,Zuev2013subset}
and directional simulation (DS) \cite{Ditlevsen1990general} among other.
However, the number of simulations required using these improvements
are still in the orders of thousands.

An alternate to the sampling based approaches discussed above is the
non-sampling based approaches.
In these methods, one first solves an optimization problem 
to determine the point on the limit-state function (a function that 
separates the failure and safe domain) that is nearest to the origin.
This operation is carried out in the standard Gaussian space and
the point obtained is often referred to as the most-probable point.
Thereafter, the limit-state function near the most-probable point
is approximated by using Taylor's series expansion and asymptotic
methods are employed to approximate the multivariate integral.
First-order reliability method (FORM) \cite{Hohenbichler1987new,Zhao1999a,Hu2015first} and second-order reliability method (SORM) \cite{Zhang2010a,Lee2012a} are the most popular non-sampling based approaches.
Different improvements to this algorithm can be found in the literature \cite{kiureghian1991efficient,koyluoglu1994new,zhao1999general}.
Non-sampling based approaches can effectively solve linear and weekly nonlinear problems.
As for computational cost, these methods are more efficient than 
the sampling based approaches and the number of simulations required is generally in the order of hundreds.

Surrogate based approaches are also quite popular for solving
reliability analysis problems.
In this method, a statistical model is used as a {\it surrogate}
to the actual limit-state function.
To train the surrogate model, input training samples are first
generated by using some design of experiment (DOE) scheme \cite{Bhattacharyya2018a,Chakraborty2016sequential}.
Responses corresponding to the training samples are generated by
simulating the true limit-state function.
Finally, some loss-function along with the training data set is used to
train the surrogate model.
Popular surrogate models available in the literature includes 
polynomial chaos expansion \cite{Sudret2008global,Xiu2002the}, analysis-of-variance decomposition \cite{Yang2012adaptive,Chakraborty2017towards,Chakraborty2016modelling},
Gaussian process \cite{Bilionis2012multi,Chakraborty2019graph,Bilionis2013multi,Nayek2019a}, artificial neural networks \cite{HosniElhewy2006reliability,Hurtado2001neural} and support vector machine \cite{Dai2012structural,Guo2009application,roy2020support,ghosh2018support} among others.
Use of hybrid surrogate models \cite{schobi2015polynomial,chakraborty2017efficient,kersaudy2015new,chakraborty2017moment,goswami2019threshold,schobi2017rare,chakraborty2017hybrid}, a surrogate model that combines more 
than one surrogate model, is also popular in the reliability analysis 
community.
Accuracy of surrogate based approaches resides somewhere between the 
sampling and non-sampling based approaches.
The computational cost of surrogate based approaches is governed by
the number of training samples required; 
this can vary from tens to thousands depending on the
nonlinearity and dimensionality of the problem.

Based on the discussion above, it is safe to conclude that 
the primary bottleneck of all the
reliability analysis techniques is the need for running simulation
to evaluate the limit-state function.
Often, the limit-state function are in form of complex nonlinear 
ordinary/partial differential equations (ODE/PDE) and solving it repeatedly
can make the process computationally expensive.
In this work, a {\it simulation free} method is proposed for solving
reliability analysis problems.
The proposed approach is rooted in a recently developed deep learning
method, referred to as the physics-informed neural network (PINN) \cite{goswami2020transfer,raissi2019physics,zhu2019physics}.
This framework requires {\it no simulation data}; instead, the 
deep neural network model is directly trained by using the physics of
the problem defined using some ODE/PDE.
For formulating a physics-informed loss-function, 
one of the recent path breaking 
developments, {\it automatic differentiation} \cite{baydin2017automatic} is used.
Using physics-informed loss-function also ensure that 
all the symmetries, invariances and 
conservation law associated with the problem are satisfied in an
approximate manner \cite{raissi2019physics}.
It is expected that this paper will lay foundation for a new paradigm 
in reliability analysis.

The rest of the paper is organised as follows.
The general problem setup is presented in \autoref{sec:ps}.
The proposed simulation free reliability analysis framework is presented in \autoref{sec:sf_ra}.
Applicability of the proposed approach is illustrated in \autoref{sec:ni} with three reliability analysis problems.
Finally, \autoref{sec:conclusions} presents the concluding remarks and future directions.
\section{Problem statement}
\label{sec:ps}
Consider an $N-$dimensional stochastic input, $\bm \Xi = \left(\Xi_1,\ldots, \Xi_N \right): \Omega_{\bm \Xi} \rightarrow \mathbb R^N$ with cumulative
distribution function $F_{\bm \Xi}(\bm \xi) = \mathbb P \left(\bm \Xi \le \bm \xi \right)$ where
$\mathbb P (\bullet)$ represents probability, $\Omega_{\bm \Xi}$ is the problem 
domain and $\bm \xi$ is a realization of
the stochastic variable $\bm \Xi$.
Now, assuming $\mathcal J \left( \bm \xi \right) = 0$ to be the limit-state function
and $\Omega_{\bm \Xi}^F \triangleq \left\{ \bm \xi: \mathcal{J}(\bm \xi) < 0\right\}$ to be the failure domain,
the probability of failure is defined as
\begin{equation}\label{eq:pf}
\begin{split}
    P_f = \mathbb P (\bm \xi \in \Omega_{\bm \Xi}^F) & = \int_{\Omega_{\bm \Xi}^F} \text d F_{\bm \Xi}(\bm \xi)\\
    & = \int_{\Omega_{\bm \Xi}} \mathbb I_{\Omega_{\bm \Xi}^F} \left( \bm \xi\right) \text d F_{\bm \Xi}(\bm \xi),
\end{split}
\end{equation}
where $\mathbb I_k (\bm \xi)$ is an indicator function such that
\begin{equation}
    \mathbb I_k (\bm \xi) = \left\{\begin{array}{l}
         1\;\text{if }\bm \xi \in k  \\
         0\;\text{elsewhere} 
    \end{array}\right. .
\end{equation}
Clearly, the limit-state function $\mathcal J (\bm \xi)$ plays a vital role in reliability
analysis.
Often, the limit-state function is in form 
of a ODE or PDE, and for computing the 
probability of failure, one needs to repeatedly
solve it.
The objective of this study is to develop a 
reliability analysis method that will be able to evaluate \autoref{eq:pf} without { even}
running a single simulation.
\section{Simulation free reliability analysis framework}
\label{sec:sf_ra}
In this section, details on the proposed 
simulation free reliability analysis framework 
is furnished.
However, before providing the details on the
proposed framework, the fundamentals of 
neural networks and its transition from
data-driven to physics-informed is presented.
The physics-informed neural network is the backbone of the proposed 
simulation free reliability analysis framework.
\subsection{A primer on neural networks}
\label{subsec:nn}
\begin{figure}[ht!]
    \centering
    \subfigure[Shallow neural network]{
    \includegraphics[width=0.4\textwidth]{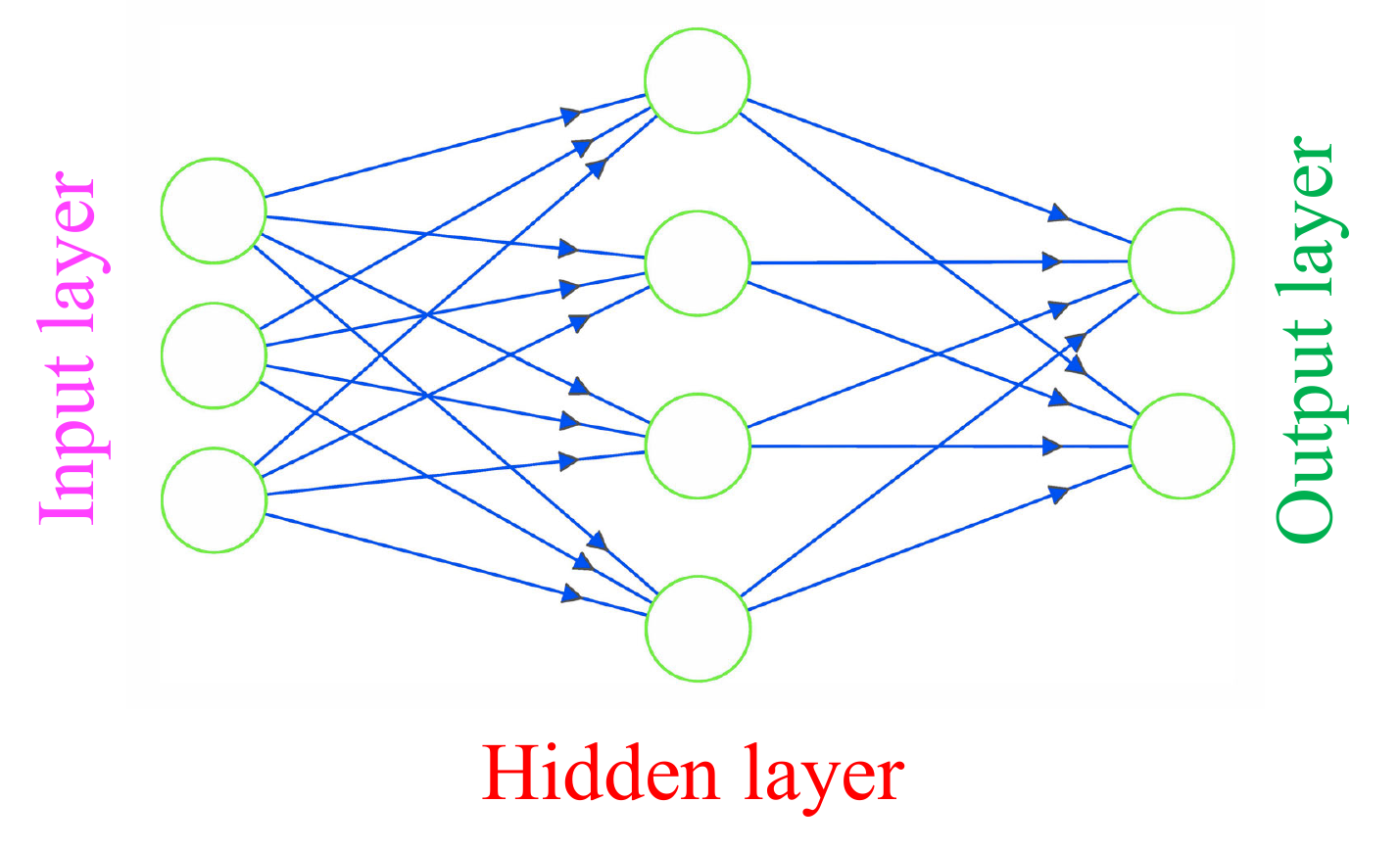}}
    \subfigure[Deep neural network]{
    \includegraphics[width=0.55\textwidth]{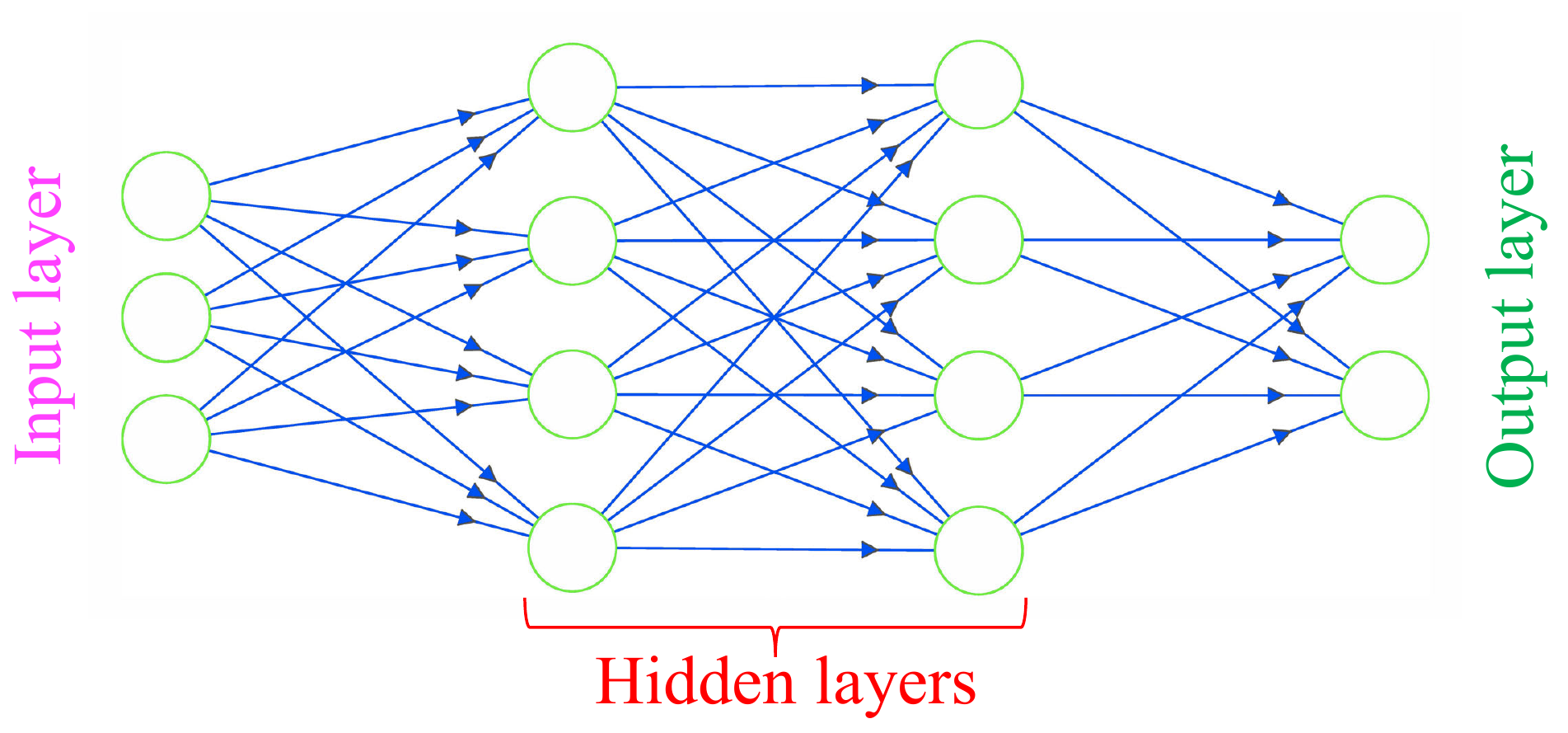}}
    \caption{Schematic representation of deep and shallow neural networks}
    \label{fig:ann}
\end{figure}

Artificial neural networks, or ANN are 
a class powerful machine learning tools
that are often used for solving regression and 
classification problems.
The idea of ANN is vaguely inspired from
an human brain. 
It consist of a set of nodes
and edges.
ANN performs a nonlinear mapping and hence,
has more expressive capability.
In theory, ANN can approximate any 
continuous function within a
given range \cite{goodfellow2016deep}.
However, in practice, an ANN often needs 
a large amount of data to 
actually learn a meaningful mapping 
between the inputs and the output.
A schematic representation of an ANN, with its
different components is shown in \autoref{fig:ann}.
Of late, `deep neural networks' (DNN) -- an ANN having more than one hidden layer, have become 
popular.
The idea is, with more hidden layers, the neural
network will be able to capture the input-output
mapping more accurately.
DNN and its associated techniques (for learning the parameters) are also referred
to as `deep learning' (DL).
In this work, fully connected deep neural network (FC-DNN) has been used; therefore, the discussion hereafter is mostly focused on FC-DNN.

Assuming a neural network with $L$ hidden layers,
the weighted input into the $i$-th neuron on
layer $l$ is represented as
\begin{equation}\label{eq:nn1}
    z_i^l = \sigma _{l-1}\left[ \sum_{k=1}^{n_{l-1}} \left(W_{i,k}^{l} \cdot z_k^{l-1}  + b_i^{l} \right) \right],
\end{equation}
where $W_{i,k}^{l}$ and $b_i^{l}$, respectively represent the weights and biases for the selected neuron.
$\sigma _{l-1} \left( \bullet \right)$ in \autoref{eq:nn1} represents the non-linear activation function.
It is assumed that the layer $(l-1)$ has $n_{l-1}$ neurons.
Note that in the above representation, it is assumed that the $0$-th layer is the input
layer and $\left(L+1\right)$-th layer is the output 
layer.
Using \autoref{eq:nn1}, the output response
in DNN is represented as
\begin{equation}\label{eq:feedforward}
\begin{split}
       \bm{Y}^L &= \sigma_L(\mathbf W^{L+1}\bm{z}^L + \bm b^L),\\
       \bm{z}^L & = \sigma_{L-1}\left(\mathbf{W}^{L}\bm{z}^{L-1} + \bm b^L\right),\\
       \bm{z}^{L-1} & = \sigma_{L-2}\left(\mathbf{W}^{L-1}\bm{z}^{L-2} + \bm b^{L-1}\right),\\
        &\vdots\\
        \bm z^1 &= \sigma_0\left(\mathbf W^1\bm{\Xi} + \bm b^1\right).\\
\end{split}
\end{equation}
$\bm Y$ in \autoref{eq:feedforward} represents the output response. 
For ease of representation, \autoref{eq:feedforward} can be represented in a
more compact form as
\begin{equation}
    \bm Y = \mathbb N \left(\bm \Xi ; \bm \theta \right),
\end{equation}
where $\mathbb N \left( \bullet ; \bm \theta \right)$ represents the neural network
operator with parameters $\bm \theta = \left[\mathbf W, \bm b  \right]$.
For utilizing a neural network in practice,
the parameters $\bm \theta$ of the neural network needs to be tuned.
This is achieved by minimizing
some loss-function that
force the neural network output to closely match the collected
data, $\mathcal D = \left\{ \bm \Xi_i, \bm Y_i \right\}_{i=1}^{N_s}$.
In literature, there exists a plethora of 
loss-functions. 
Interested readers can refer \cite{goodfellow2016deep,beale1996neural} to get an account of different loss-functions available
in the literature.
Since, the DNN discussed above is dependent on
data, $\mathcal D$, it is referred to as the
{\it data-driven} DNN.
\subsection{From data-driven to physics-informed DNN}
\label{subsec:pidnn}
Over the years, {\it data-driven} DNN has 
successfully solved a wide range of problems 
spanning across various domains \cite{xin2018machine,badjatiya2017deep,carrio2017review,wang2020optimization,waring2020automated,lamy2019explainable}.
Despite such success, one major bottleneck 
of DNN is its dependence on data; this is
particularly true when the data generation
process is computationally expensive.
To address this issue, the concept of 
physics-informed deep learning was proposed
in \cite{raissi2019physics}.
Within this framework, prior knowledge about
the model, often available in form of ODE/PDE
is utilized to train a DNN model.
It was illustrated that this model can solve
complex non-linear PDEs.

Consider a nonlinear PDE of the form
\begin{equation}\label{eq:pde}
    u_t + g\left[u,u_x,u_{xx},\ldots ;\lambda  \right] = 0,
\end{equation}
where $u$ is the unknown solution and 
$g\left[\bullet;\lambda\right]$ is a nonlinear
operator parameterized by $\lambda$.
The subscripts in \autoref{eq:pde} represents 
derivative with respect to space and/or time.
\begin{equation}
        u_t = \frac{\partial u}{\partial t}, \;
        u_x = \frac{\partial  u}{\partial  x}, \;
        u_{xx} = \frac{\partial^2 u}{\partial x^2}, \;
        \cdots
\end{equation}
In physics-informed deep learning,
the objective is to use \autoref{eq:pde} to 
train a deep neural network model that 
approximates the unknown variable $u$ in
\autoref{eq:pde}.
This is achieved by representing the unknown response $u$ using a neural network
\begin{equation}\label{eq:step1_dnn}
    u(x,t) \approx \hat u(x,t) = \mathbb N\left(x,t;\bm \theta \right),
\end{equation}
and following four simple steps
\begin{itemize}
    \item Generate a set of collocation points $\mathcal D_c = \left\{ x_i, t_i \right\}_{i=1}^{N_c}$, where $N_c$ is the number of collocation points generated.
    Also generate input data corresponding to the boundary and initial conditions, $\mathcal D_b = \left\{x_b, t_i \right\}_{i=1}^{N_b}$ and 
    $\mathcal D_i = \left\{x_i, t_0 \right\}_{i=1}^{N_i}$. $N_b$ and $N_i$ respectively represents the number of 
    points corresponding to boundary condition and initial condition. $t_0$ is the initial time and $x_b$ is the coordinate where the boundary condition is imposed.
    \item Based on the generated collocation
    points, a {\it physics-informed} loss-function is formulated as
    \begin{equation}\label{eq:pinn_s1}
        \mathcal{L}_p = \frac{1}{N_c}\sum_{i=1}^{N_c}\left( R(x_i,t_i)\right)^2,
    \end{equation}
    where
    \begin{equation}\label{eq:pinn_s1a}
        R(x_i,t_i) = \hat u_t(x_i, t_i) + g\left( \hat u(x_i, t_i), \hat u_x((x_i, t_i), \hat u_{xx}(x_i, t_i), \ldots; \lambda \right)
    \end{equation}
    is the residual of the PDE calculated at $(x_i, t_i)$.  
    $\hat u$ in \autoref{eq:pinn_s1a} is a DNN and is parameterized 
    by $\bm \theta$. The derivatives of $\hat u_t(x_i, t_i)$, i.e, $\hat u_x(x_i, t_i)$, $\hat u_{xx}(x_i, t_i)$ etc are calculated by using AD.
    \item For the boundary and initial conditions,
    formulate a data-driven loss-functions
    \begin{subequations}\label{eq:loss_data}
    \begin{equation}\label{eq:loss_b}
        \mathcal L_b = \frac{1}{N_b}\sum_{k=1}^{N_b}\left[ u_{b,k} - \hat u(x_k, t_k) \right]^2,
    \end{equation}
    \begin{equation}\label{eq:loss_i}
        \mathcal L_i = \frac{1}{N_i}\sum_{k=1}^{N_i}\left[ u_{i,k} - \hat u(x_k, t_k) \right]^2.
    \end{equation}
    \end{subequations}
    $u_{i,k}$ and $u_{b,k}$ in \autoref{eq:loss_data} represent the initial and boundary conditions of the problem. $\hat u$ as before is a DNN.
    \item Formulate the combined loss-function, $\mathcal L$
    \begin{equation}
        \mathcal L = \mathcal L_p + \mathcal L_b + \mathcal L_i,
    \end{equation}
    and minimize it to obtain the parameters, $\bm \theta$.
\end{itemize}
Once the model is trained, it is used, as usual, to make predictions at unknown point ${x^*,t^*}$.
\subsection{Proposed approach}
\label{subsec:pa}
In this section, 
the physics-informed DNN presented
in \autoref{subsec:pidnn} is extended for solving 
reliability analysis problems.
Consider, the limit-state function $\mathcal J(\bm \xi )$ discussed in \autoref{sec:ps} is represented as
\begin{equation}
    \mathcal{J}(\xi) = u(\xi) - u_0,
\end{equation}
where $u(\xi)$ is the response and $u_0$ indicates the threshold value.
Also assume that $u(\xi)$
is obtained by solving a stochastic PDE of the form
\begin{equation}\label{eq:spde}
    u_t + g\left[u,u_x,u_{xx},\ldots ;\xi  \right] = 0.
\end{equation}
Note that \autoref{eq:spde} is assumed to
have the same functional form as \autoref{eq:pde}; the only difference is that
the parameter $\lambda$ is replaced 
with $\xi$.
This indicates that parameters are considered to be stochastic.
With this setup, the objective now is to 
train a DNN that can act as a surrogate to
the response $u$.
In a conventional data-driven setup, one first
generate training samples from the stochastic input, $\xi_i,\,i=1,\ldots,N_s$,
runs a PDE solver such as finite element (FE) method $N_s$ times
to generate outputs, $u_i,\,i=1,\ldots,N_s$ and then train a data-driven DNN model.
Because of the need to run $N_s$ FE simulations,
such a data-driven approach can quickly become 
computationally prohibitive for systems defined
by complex nonlinear PDEs.
This paper takes a different route and 
attempts to directly develop a DNN based 
surrogate from the stochastic PDE in \autoref{eq:spde}.
To that end, the stochastic response is first
represented in form of a DNN
\begin{equation}
    u(x,t,\xi) \approx u_{NN}(x,t,\xi) = \mathbb N \left(x,t,\xi; \bm \theta \right).
\end{equation}
Note that unlike \autoref{eq:step1_dnn}, the
input to the neural network now also includes the system parameters, $\xi$.
Next, the neural network output is modified to automatically satisfy the initial and/or boundary conditions,
\begin{equation}\label{eq:bs_pidnn}
    \hat u \left(x,t,\xi \right) = u_{b,i}(x_b,t_i) + B \cdot u_{NN}(x,t,\xi).
\end{equation}
The function $B$ is defined in such as way so 
that $B=0$ at the boundary and initial points.
The function $u_{b,i}(x_b,t_i)$, on the other hand, is defined using the initial and boundary 
conditions of the problem.
For example, if the boundary condition demands at $x=0$, $u=0$ and the initial condition demands at $t=0$, $u=0$, one sets
\begin{equation}
    \hat u(x,t,\xi) = x\cdot t \cdot u_{NN}(c,t,\xi).
\end{equation}
More examples on how the initial and boundary conditions are automatically satisfied are provided in \autoref{sec:ni}.
Note that $\hat u(x,t,\xi)$ in \autoref{eq:bs_pidnn} can also be viewed as 
a neural network, $\hat{\mathbb N}(x,t,\xi; \bm \theta)$ with the same parameters $\bm \theta$.
The derivatives present in PDE are calculated
from $\hat{\mathbb N}(x,t,\xi; \bm \theta)$,
by using AD.
\begin{equation}\label{eq:AD}
\begin{split}
    u_t \approx \frac{\partial \hat u}{\partial t} & = \hat{\mathbb N}^t(x,t,\xi;\bm \theta),\\
    u_x \approx \frac{\partial \hat u}{\partial x} & = \hat{\mathbb N}^x(x,t,\xi;\bm \theta),\\
    u_{xx} \approx \frac{\partial^2 \hat u}{\partial x^2} & = \hat{\mathbb N}^{xx}(x,t,\xi;\bm \theta),\\
    & \vdots
\end{split}
\end{equation}
Note that all the derivatives computed
in \autoref{eq:AD} are deep neural networks with the same exact architecture and parameters and hence, have been denoted using $\hat{\mathbb N}^t(\cdot,\cdot,\cdot;\bm\theta)$, 
$\hat{\mathbb N}^x(\cdot,\cdot,\cdot;\bm\theta)$ and
$\hat{\mathbb N}^xx(\cdot,\cdot,\cdot;\bm\theta)$.
The only difference between the original DNN,
$\hat{\mathbb N}(\cdot,\cdot,\cdot;\bm\theta)$
and those derived in \autoref{eq:AD} resides in
the form of the activation function.
Using DNNs, the residual of the PDE is defined
as
\begin{equation}\label{eq:res_dnn}
    R = \hat{\mathbb N}^t(x,t,\xi;\bm\theta) + g\left[\hat{\mathbb N}(x,t,\xi;\bm\theta),\hat{\mathbb N}^x(x,t,\xi;\bm\theta),\hat{\mathbb N}^{xx}(x,t,\xi;\bm\theta),\ldots \right] = \hat{\mathbb N}^R(x,t,\xi;\bm\theta) 
\end{equation}
Again, the operations carried out in \autoref{eq:res_dnn} yields a DNN 
$\hat{\mathbb N}^R(\cdot,\cdot,\cdot;\bm\theta)$ having
the same parameters $\bm \theta$.
When trained, $\hat{\mathbb N}^R(\cdot,\cdot,\cdot;\bm\theta)$
ensures that the stochastic PDE is satisfied
and hence,
is referred to as the physics-informed DNN.
In the ideal scenario, 
$\hat{\mathbb N}^R(x,t,\xi;\bm\theta) =0: \xi \in \Omega_{\xi}, x \in \Omega_x, t\in\Omega_t$.

\begin{figure}[ht!]
    \centering
    \includegraphics[width = 0.9\textwidth]{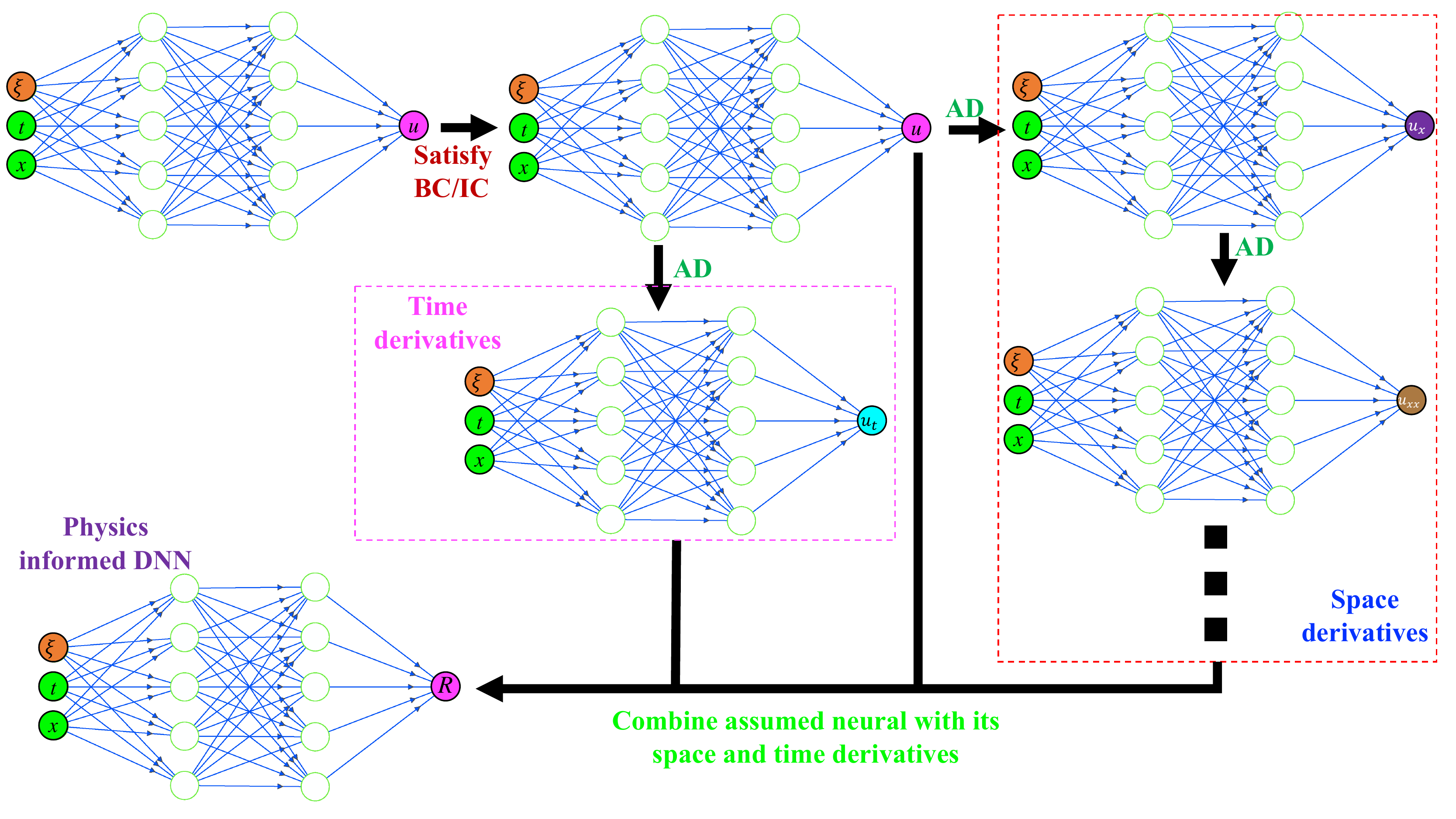}
    \caption{Proposed physics-informed DNN based framework for simulation free reliability analysis. Along with the spatial and temporal coordinates, the stochastic parameter is also an input to the DNN. All the required derivatives are taken by using automatic differentiation.}
    \label{fig:flowchart}
\end{figure}

To train the network and compute the parameters,
$\bm \theta$, three simple steps are followed.
\begin{itemize}
    \item Generate collocation points, $\mathcal D = \left\{x_i, t_i, \xi_i \right\}_{i=1}^{N_c}$ 
    are generated by using some suitable DOE scheme.
    \item Formulate the loss-function as
    \begin{equation}
        \mathcal L(\bm \theta) = \frac{1}{N_c}\sum_{k=1}^{N_c}\left[ \hat{\mathbb N}^R(x_k,t_k,\xi_k;\bm\theta) ^2\right].
    \end{equation}
    \item Compute $\bm \theta$ by minimizing the loss-function
    \begin{equation}
        \bm \theta ^* = \arg\min_{\bm \theta} \mathcal L(\bm \theta).
    \end{equation}
\end{itemize}
Once the parameters $\bm \theta$ has been estimated, $\hat{\mathbb N}(x,t,\xi; \bm \theta)$ (i.e., \autoref{eq:bs_pidnn}) can
be used to predict response at any unknown point $(x^*, t^*, \xi^*)$.
The framework proposed is coded using \texttt{TensorFlow} \cite{abadi2016tensorflow}.
For minimizing the loss-function, RMSprop optimizer \cite{tieleman2012lecture} followed by L-BFGS has been used.
Details on parameters of the optimizers are
provided in \autoref{sec:ni}.
A schematic representation of the proposed framework is shown in \autoref{fig:flowchart}.

The proposed approach has a number of advantages.
\begin{itemize}
    \item The primary bottleneck in reliability analysis is the need for running computationally expensive simulations.
    The proposed eliminates this bottleneck as it needs {\it no simulations}.
    \item Since the proposed approach trains the
    DNN model from the governing PDE/ODE, the 
    solution obtained using the trained model
    satisfies physical laws such as invariances and conservation laws.
    \item Unlike most of the methods, the 
    proposed approach provides a continuous solution; the response $u$ can be evaluated
    at any given point in time and space.
    This will be extremely helpful in solving 
    time-dependent reliability analysis problems.
\end{itemize}
\section{Numerical illustration}\label{sec:ni}
Numerical examples are presented in this section
to illustrate the performance of the proposed 
approach.
The examples selected involve a wide variety of
problems with limit-state functions
defined using ODE and PDE, single equation and system of equations,
linear and nonlinear ODE/PDE, and also 
univariate and multivariate problems.
The computational complexity of the problems
selected are not high; this enables generating
solution using MCS and other state-of-the-art 
reliability analysis methods for comparison.
The software accompanying the proposed approach
is developed in \texttt{python} using \texttt{TensorFlow} \cite{abadi2016tensorflow}.
For generating the benchmark solutions, \texttt{MATLAB} \cite{matlab2019} has been used.
\subsection{Ordinary differential equation}\label{subsec:ode}
As the first example, a simple stochastic ODE is considered.
\begin{equation}\label{eq:eg1_ps}
    \frac{\text d u}{\text d t} = -Z u,
\end{equation}
where $Z$ is the decay rate coefficient and is
considered to be stochastic.
The ODE in \autoref{eq:eg1_ps} is subjected to the following initial condition,
\begin{equation}\label{eq:eg1_ic}
    u(t=0) = u_0,
\end{equation}
The exact solution for this problem is known.
\begin{equation}
    u\left(t,Z\right) = u_0\exp\left(-Zt\right).
\end{equation}
For reliability analysis, a limit-state function
is defined as
\begin{equation}\label{eq:eg1_ls}
    \mathcal J(u\left(t,Z\right)) = u\left(t,Z\right) - u_d,
\end{equation}
where $u_d$ is the threshold value.
This problem has previously been studied in \cite{li2010evaluation}.

For this particular example, the stochastic variable $Z \sim \mathcal N \left(\mu, \sigma^2 \right)$ is considered to be following normal distribution with mean, $\mu = -2$ and the 
standard deviation, $\sigma = 1$.
The threshold value, $u_d = 0.5$, and the initial value, $u_0 = 1.0$ is considered.
The exact failure probability for this problem
is $P_f = 0.003539$.
MCS with $10^6$ simulations yields a failure
probability of $P_f = 0.03496$.

For solving the problem using the proposed 
simulation free reliability analysis framework,
the unknown $u$ is represented using a FC-DNN with 2 hidden layers, with each hidden layer
having 50 neurons.
The DNN has 2 inputs, time $t$ and the stochastic parameter $Z$.
The activation function for all but the last
layer is considered to be a hyperbolic tangent 
function (\texttt{tanh}).
For the last layer, a linear activation is used.
To automatically satisfy the initial condition,
the DNN output is modified as
\begin{equation}\label{eq:eg1_dnn_bc}
    \hat u = t\cdot u_{NN} + u_0,
\end{equation}
where $u_0 = 1.0$ and $u_{NN}$
is the DNN output.
The residual for formulating the loss-function
is represented as
\begin{equation}\label{eq:eg1_res}
    R_i = \frac{\text d \hat u}{\text d t} + Z \hat u,
\end{equation}
where $\hat u $ is obtained from \autoref{eq:eg1_dnn_bc}.
For training the model, 4000 collocation points
have been generated using the Latin hypercube sampling \cite{Iman1980latin}.
Th RMSprop optimizer is run for 10,000 iterations.
The maximum allowable iterations for the L-BFGS optimizer is set to 50,000.

The results obtained using the proposed approach
and MCS are shown in \autoref{tab:eg1_res}.
Reliability index $\beta$
is also computed,
\begin{equation}\label{eq:ri}
    \beta = \Phi^{-1}(1- P_f),
\end{equation}
where $\Phi(\bullet)$ is the cumulative distribution function of standard normal
distribution.
The results obtained using the proposed approach
is found to closely match the exact solutions and MCS results.
For sake of comparison, results using FORM, SORM, IS, SS and DS have also been reported in \autoref{tab:eg1_res}.
All the other methods are found to
yield results with similar 
kind of accuracy.
As for efficiency, FORM, SORM, IS, DS and SS requires 42, 44, 1000, 7833 and 1199 simulations, respectively.
The proposed approach, on the other hand, needs
no simulation for generating the probability of
failure estimates. 

\begin{table}[ht!]
    \caption{Reliability analysis results for the stochastic ODE problem using various methods.}
    \label{tab:eg1_res}
    \centering
    \begin{tabular}{lcccc}
    \hline
    \textbf{Methods} & $P_f$ & $\beta$ & $N_s$ & $\epsilon = \frac{\left| \beta_e - \beta \right|}{\beta_e}$ \\ \hline
    Exact & $0.003539$ & $2.6932$ & -- & -- \\ \hline \hline
     MCS & $0.0035$ & $2.6949$ & $10^6$ & 0.06\% \\
     FORM & $0.0036$ & $2.6874$ & $42$ & 0.21\% \\
     SORM & $0.0036$ & $2.6874$ & $44$ & 0.21\% \\
     IS & 0.0034 & $2.7074$ & $1000$ & 0.52\% \\
     DS & 0.0034 & $2.7074$ & $7833$ & 0.52\% \\
     SS & $0.0030$ & $2.7456$ & $1199$ & 1.95\% \\
     PI-DNN & $0.0035$ & $2.6949$ & $0$ & 0.06\% \\ \hline
    \end{tabular}
\end{table}

To further analyze the performance of the 
proposed method, systematic case studies
by varying the number of neurons, number 
of hidden layers and the number of collocation points have been carried out.
\autoref{tab:eg1_cs1} reports the reliability index and probability failures obtained corresponding to different 
settings of the PI-DNN.
For this particular problem, the effect of 
number of hidden layers and number of neurons 
is relatively less; results corresponding to all the settings are found to be accurate.
\begin{table}[ht!]
    \caption{Variation in reliability analysis results due to change in the DNN architecture (number of hidden layers and neurons). The numbers in the bracket indicates reliability indices.}
    \label{tab:eg1_cs1}
    \centering
    \begin{tabular}{l||cccc}
    \hline
    \backslashbox{Layers}{Neurons}   & 30 & 40 & 50 & 60  \\ \hline \hline
    2 & 0.00344 (2.7026) & 0.0036 (2.6874) & 0.00352 (2.6949) & 0.00359 (2.6884)\\
    3 & 0.0037 (2.6783)  & 0.0036 (2.6874) & 0.0036 (2.6874) & 0.0036 (2.6874) \\
    3 & 0.00357 (2.6902) & 0.00364 (2.6838) & 0.0036 (2.6874) & 0.0036 (2.6874) \\ \hline
    \end{tabular}
\end{table}
\autoref{tab:eg1_cs2} shows the variation in the probability of failure and reliability index estimates with change in the number of collocation points, $N_c$.
It is observed that the results obtained are more or less constant beyond 500 collocation points.
For 250 collocation points, the results 
obtained are found to be less accurate.
Nonetheless, it is safe to conclude that for
this problem, results obtained using the 
proposed PI-DNN based simulation free method is 
highly accurate.
\begin{table}[ht!]
    \caption{Reliability analysis results with change in number of collocation points.}
    \label{tab:eg1_cs2}
    \centering
    \begin{tabular}{l||cccccccc}
    \hline
        $N_c$ & 250 & 500 & 1000 & 2000 & 4000 & 6000 & 8000 & 10000 \\ \hline
         $P_f$ & 0.00394 & 0.0035 & 0.0035 & 0.0035 & 0.00352 & 0.00352 & 0.00352 & 0.00352 \\
         $\beta$ & 2.6572 & 2.6968 & 2.6968 & 2.6968 & 2.6949 & 2.6949 & 2.6949 & 2.6949 \\\hline
    \end{tabular}
\end{table}
\subsection{Viscous Burger's equation}\label{subsec:burger}
As the second example,  
viscous Burger's equation is considered.
The PDE of the Burger's equation is given as
\begin{equation}\label{eq:eg2_pde}
    u_t + u u_x = \nu u_{xx}, \;\;\; x\in [-1,1],
\end{equation}
where $\nu>0$ is the viscosity of the system.
The Burger's equation in \autoref{eq:eg2_pde}
is subjected to the following boundary
conditions
\begin{equation}\label{eq:eg2_bc}
    u(x=-1) = 1 + \delta\;\;u(x=1) = -1.
\end{equation}
The initial condition of the system is obtained
by linear interpolation of the boundary 
conditions.
\begin{equation}\label{eq:eg2_ic}
    u(t=0,x) = -1 + \left(1-x\right)\left(1 + \frac{\delta}{2} \right).
\end{equation}
$\delta$ in Eqs. \ref{eq:eg2_bc} and \ref{eq:eg2_ic} denotes a small perturbation
that is applied to the boundary at $x = -1$.
Solution of \autoref{eq:eg2_pde} has a transition layer at distance $z$ such that, 
$u(z) = 0$.
As already established in a number of previous 
studies \cite{xiu2004supersensitivity,lorenz1981nonlinear}, the location of the transition layer $z$
is super sensitive to the perturbation $\delta$.
A detailed study on properties of this transition layer can be found in \cite{xiu2004supersensitivity}.

In this paper, the perturbation $\delta$ is considered to be a uniformly distributed variable, $\delta \sim \mathcal U \left(0,e\right),\; e<<1$.
With this the limit-state function is defined as
\begin{equation}\label{eq:eg2_ls}
    \mathcal J \left( z(\delta) \right) = - z(\delta) + z_0,
\end{equation}
where $z_0$ is the threshold.
For this problem, $e=0.1$ is considered.
Different case studies are performed by varying the threshold parameter, $z_0$.

For solving this problem using the proposed 
approach, the unknown variable $u$ is first
represented by using a FC-DNN.
The FC-DNN considered has 4 hidden layers with 
each layer having 50 neurons.
The DNN has three inputs, the spatial coordinate $x$, the temporal coordinate $t$ ad the stochastic variable $\delta$.
Similar to previous example, the activation function of all but the last layer is considered
to be \texttt{tanh}, and for the last layer, a linear activation function is considered.
To automatically satisfy the boundary and the 
initial condition, the neural network output 
$u_{NN}$ is modified as follows.
\begin{equation}\label{eq:eg2_mod_nn}
    \hat u = (1 - x)(1+x)tu_{NN} + u(t=0,x),
\end{equation}
where $u(t=0,x)$ is defined in \autoref{eq:eg2_ic}.
The residual for formulating the physics-informed loss-function
is defined as
\begin{equation}\label{eq:eg2_res}
    R_i = \left. u_t\right|_{x=x_i,t=t_i,\delta = \delta_i} + \left. uu_x\right|_{x=x_i,t=t_i,\delta = \delta_i} -  \left. \nu  u_{xx}\right|_{x=x_i,t=t_i,\delta = \delta_i}. 
\end{equation}
For training the network, 30,000 collocation points have been generated using the Latin hypercube sampling.
The RMSProp optimizer is run for 15,000 iterations.
The maximum number of allowed iterations for 
the L-BFGS optimizer is set at 50,000.

For generating benchmark solutions, 
the deterministic Burger's equation is solved using FE in  \texttt{FENICS} \cite{alnaes2015fenics}.
The \texttt{FENICS} based solver is then 
coupled with MATLAB based \texttt{FERUM} software \cite{bourinet2009review} for generating the benchmark
solutions.
Note that the proposed PI-DNN based approach needs no simulation data and hence, no such solver is needed; instead, the PI-DNN 
is directly trained based on the physics
of the problem.

The results obtained using the proposed
approach and other state-of-the art methods
from the literature are shown in \autoref{tab:eg2_res}.
The threshold value $z_0$ is set to be 0.45
and the reliability analysis is carried out at $t = 10$.
It is observed that the proposed approach yields highly accurate results, almost matching
with the MCS solution.
Results obtained using the other methods are slightly less accurate.
As for computational efficiency, the MCS results
are obtained by running the FE solver 10,000 times. 
IS, DS, SS, FORM and SORM, respectively needs
1000, 4001, 1000, 58 and 60 runs of the FE solver.
The proposed approach, on the other hand, 
needs no simulations.

\begin{table}[ht!]
    \caption{Reliability analysis results for viscous Burger's problem.}
    \label{tab:eg2_res}
    \centering
    \begin{tabular}{lcccc}
    \hline
    \textbf{Methods} & $P_f$ & $\beta$ & $N_s$ & $\epsilon = \frac{\left| \beta_e - \beta \right|}{\beta_e}$  \\ \hline
    MCS & 0.1037 & 1.2607  & 10,000 & -- \\ \hline \hline 
    FORM & 0.1091 & 1.2313 & 58 & 2.33\%  \\
    SORM & 0.1091 & 1.2313 & 60 & 2.33\% \\
    IS & 0.1126 & 1.2128 & 1000 & 3.80\% \\
    DS & 0.0653 & 1.5117 & 4001 & 19.9\% \\
    SS & 0.0800 & 1.4051 & 1000 & 11.45\% \\
    PI-DNN & 0.0999 & 1.2821 & 0 & 1.70 \% \\ \hline 
    \end{tabular}
\end{table}

\begin{figure}[ht!]
    \centering
    \includegraphics[width = 0.6\textwidth]{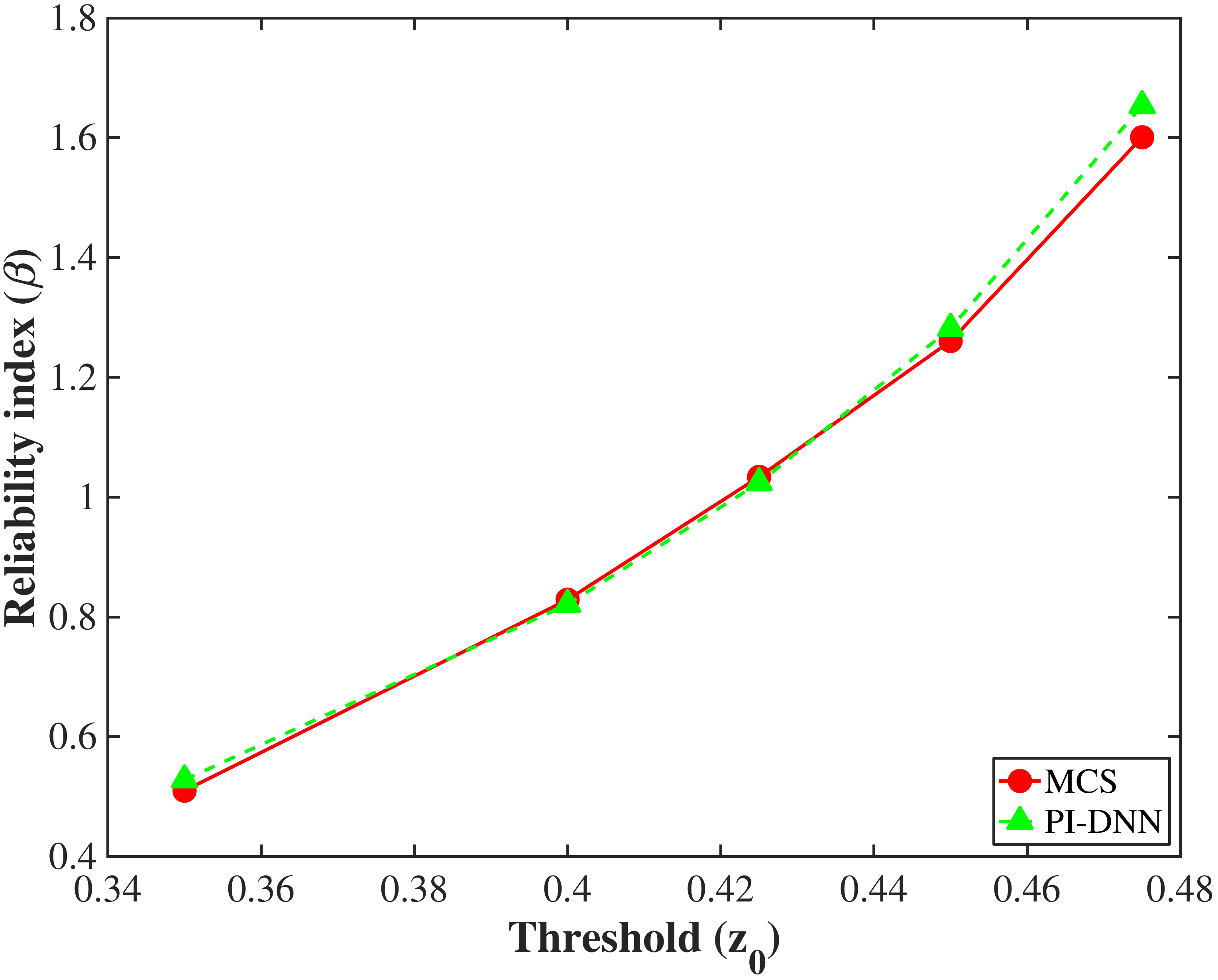}
    \caption{Reliability index corresponding to different thresholds $z_0$ using MCS and PI-DNN.}
    \label{fig:eg2_thres_vary}
\end{figure}

Next, the performance of the proposed PI-DNN 
in predicting the reliability corresponding to
different thresholds is examined.
Through this, it is possible to examine whether
the proposed approach is able to emulate the
FE solver properly.
The results obtained are shown in \autoref{fig:eg2_thres_vary}.
Results corresponding to MCS are also shown.
It is observed that the proposed approach 
is able to predict the reliability index corresponding to all the thresholds.
This indicates that the DNN, trained only
from the governing PDE, is able to reasonably 
emulate the actual FE solver.

Lastly, the effect of network architecture and 
number of collocation points on the performance of the proposed approach is examined.
\autoref{tab:eg2_comp1} shows the results 
corresponding to different number of collocation points.
It is observed that the results improve with 
increase in the number of collocation points.
Beyond 30,000 collocation points, the results
is found to stabilize with no significant change.
\begin{table}[ht!]
    \caption{Performance of PI-DNN with change in number of collocation points.}
    \label{tab:eg2_comp1}
    \centering
    \begin{tabular}{l||cccccc}
    \hline
    \backslashbox{$z_0$}{$N_c$} &  5000 & 15000 & 20000 & 25000 & 30000    & 35000 \\ \hline \hline
    0.45 & 0  & 0 & 0.0697 & 0.0791 & 0.0999     & 0.1001 \\
    0.40 & 0 & 0.1064 & 0.1802 & 0.1893 & 0.2058     & 0.2059 \\
    0.35 & 0.0852 & 0.2361 & 0.2858 & 0.2928 & 0.3050     & 0.3051 \\ \hline
    \end{tabular}
\end{table}
Tables \ref{tab:eg2_comp2} and \ref{tab:eg2_comp3} show the probability of failure and reliability index estimates corresponding to different number of hidden layers and neurons. It is observed that with too few layers/neurons, the DNN is unable to track the probability of failure.
On the other hand, too many neurons/layers 
results in over-fitting.
One way to address this over-fitting issue is
to use some form of regularizer in the loss-function.
However, this is not within the scope of the current work.
\begin{table}[ht!]
    \caption{Variation in reliability analysis results due to change in number of neurons and number of hidden layers. $z_0 = 0.40$ is considered. The numbers in the bracket indicates reliability index.}
    \label{tab:eg2_comp2}
    \centering
    \begin{tabular}{l||ccc}
    \hline
    \backslashbox{Layers}{Neurons} & 40 & 50 & 60 \\ \hline \hline 
    2   & 0.0 ($\infty$) &	0.0 ($\infty$) & 0.0 ($\infty$)  \\
    3   & 0.0584 (1.5683) &	0.1318 (1.1179) & 0.19 (0.8779) \\
    4   & 0.1741 (0.9380) &	0.2058 (0.9380) & 0.1721 (0.9459)  \\
    5   & 0.1918 (0.8713) &	0.1924 (0.8691) & 0.1857 (0.8939)  \\
    6   & 0.1795 (0.9173) &	0.1723 (0.9451) & 0.19 (0.8779) \\ \hline 
    \end{tabular}
\end{table}
\begin{table}[ht!]
    \caption{Variation in reliability analysis results due to change in number of neurons and number of hidden layers. $z_0 = 0.45$ is considered. The numbers in the bracket indicates reliability index}
    \label{tab:eg2_comp3}
    \centering
    \begin{tabular}{l||ccc}
    \hline
    \backslashbox{Layers}{Neurons} & 40 & 50 & 60 \\ \hline \hline 
    2   & 0.0 ($\infty$) &	0.0 ($\infty$) & 0.0 ($\infty$)  \\
    3   & 0.0 ($\infty$) &	0.0 ($\infty$) & 0.0836 (1.3813) \\
    4   & 0.0598 (1.5565) &	0.0999 (1.2821) & 0.056 (1.5893)  \\
    5   & 0.0759 (1.4332) &	0.0855 (1.3690) & 0.0741 (1.4459) \\
    6   & 0.0732 (1.4524) &	0.0587 (1.5658) & 0.0699 (1.4765)  \\ \hline 
    \end{tabular}
\end{table}
\subsection{Systems of equations: cell-signalling cascade}
\label{subsec:eg3}
As the final example of this paper, 
a mathematical model of the autocrine cell signalling cascade is considered.
This model was first proposed in \cite{shvartsman2002autocrine}.
Considering, $e_{1p}$, $e_{2p}$ and $e_{3p}$
to be the concentrations of the active form of enzymes, the governing differential equations
for this system is given as
\begin{equation}\label{eq:eg3_odes}
    \begin{split}
        \frac{\text d e_{1p}}{\text d t} & = \frac{I(t)}{1 + G_4 e_{3p}} \frac{V_{\max , 1}(1 - e_{1p})}{K_{m,1} + \left( 1 - e_{1p} \right)} - \frac{V_{\max ,2}e_{1p}}{K_{m,2} + e_{1p}}, \\
        \frac{\text d e_{2p}}{\text d t} & =  \frac{V_{\max , 3}e_{1p}(1 - e_{2p})}{K_{m,3} + \left( 1 - e_{2p} \right)} - \frac{V_{\max ,4}e_{2p}}{K_{m,4} + e_{2p}}, \\
        \frac{\text d e_{3p}}{\text d t} & =  \frac{V_{\max , 5}e_{2p}(1 - e_{3p})}{K_{m,5} + \left( 1 - e_{3p} \right)} - \frac{V_{\max ,6}e_{3p}}{K_{m,6} + e_{3p}},
    \end{split}
\end{equation}
where $G_4 = 0$, $I(t) = 1$ and $K_{m,i} = 0.2,\,\forall i$. The ODEs in \autoref{eq:eg3_odes}
are subjected to the initial condition,
\begin{equation}\label{eq:eg3_ic}
    e_{1p}(t=0) = 0,\; e_{2p}(t=0) = 1,\; e_{3p}(t=0) = 0.  
\end{equation}
The parameters $V_{\max , i}, \, i=1,\ldots, 6$ 
are considered to be stochastic.
This is a well-known benchmark problem previously studied in \cite{li2010evaluation}.

For this problem, $V_{\max , i}$ is defined as
\begin{equation}
    V_{\max , i} = \left< V \right>_{\max,i}\left(1 + \sigma Z_i \right),\; i=1,\ldots,6,
\end{equation}
where $\sigma = 0.1$.
The variable $Z_i$ accounts for the uncertainty
in $V_{\max,i}$.
It is assumed that $Z_i \sim \mathcal U(-1,1)$ is uniformly distributed with lower-limit $-1$ and upper-limit 1.
Therefore this problem has six stochastic variables.

The limit-state function for this problem is 
defined as
\begin{equation}\label{eq:eg3_ls}
    \mathcal J (Z,t) = e_{3p}(Z,t) - e_{3p,0},
\end{equation}
where $e_{3p,0}$ is the threshold value.
Similar to \autoref{subsec:burger}, 
results corresponding to to different threshold values are presented.

For solving the problem using the proposed 
PI-DNN based simulation free approach, a
FC-DINN with four hidden layers is considered.
Each hidden layer has 50 neurons.
There are seven inputs to the DNN, the six stochastic variables, $Z_i$, $=1,\ldots, 6$ and time $t$.
There are three outputs, $e_{1p}$, $e_{2p}$ and 
$e_{3p}$.
Similar to the previous two examples, the 
\texttt{tanh} activation function for all but the last layer is considered.
For the last layer, linear activation function
is used.
The neural network output are modified as follows to automatically satisfy the initial
conditions.
\begin{equation}\label{eq:eg3_icnn}
    \hat e_{1p} = t\cdot e_{1p,NN},\;\;
    \hat e_{2p} = t\cdot e_{2p,NN} + 1.0,\;\;
    \hat e_{3p} = t\cdot e_{3p,NN}
\end{equation}
where $e_{ip,NN},\,i=1,2,3$ is the raw output from the neural network.
The residuals for formulating the loss function
are defined as
\begin{equation}\label{eq:eg3_res}
    \begin{split}
        R_{1,i} & = (1 + G_4 \hat e_{3p}) (K_{m,1} + \left( 1 - \hat e_{1p} \right)) (K_{m,2} + \hat e_{1p})
        \frac{\text d \hat e_{1p}}{\text d t} - I(t)(V_{\max , 1}(1 - \hat e_{1p}))(K_{m,2} + \hat e_{1p})\\
        &              + (V_{\max ,2}\hat e_{1p})(1 + G_4 \hat e_{3p}) (K_{m,1} + \left( 1 - \hat e_{1p} \right)), \\
        R_{2,i} & = (K_{m,3} + \left( 1 - \hat e_{2p} \right)) (K_{m,4} + \hat e_{2p})
        \frac{\text d \hat e_{1p}}{\text d t} - (V_{\max , 3}\hat e_{2p}(1 - \hat e_{2p}))(K_{m,4} + \hat e_{2p})\\
        &              + (V_{\max ,4}\hat e_{2p})(K_{m,3} + \left( 1 - \hat e_{2p} \right)), \\
        R_{3,i} & = (K_{m,5} + \left( 1 - \hat e_{3p} \right)) (K_{m,6} + \hat e_{3p})
        \frac{\text d \hat e_{2p}}{\text d t} - (V_{\max , 5}\hat e_{2p}(1 - \hat e_{3p}))(K_{m,6} + \hat e_{3p})\\
        &              + (V_{\max ,6}\hat e_{3p})(K_{m,5} + \left( 1 - \hat e_{3p} \right)) .\\
    \end{split}
\end{equation}
The dependence of the residuals and the DNN on the collocation points have been removed from
brevity of representation.
The functional form of residuals presented 
in \autoref{eq:eg3_res} are obtained by carrying
out some trivial algebraic operation on the 
governing equations in \autoref{eq:eg3_odes};
this is necessary to stop the PI-DNN weights
from exploding during the training phase.
Using the residuals in \autoref{eq:eg3_res},
the physics-informed loss function for 
training the PI-DNN is represented as
\begin{equation}
    \mathcal L = \frac{1}{N_c}\sum_{k=1}^{N_c}\left( R_{1,k}^2 \right) + \frac{1}{N_c}\sum_{k=1}^{N_c}\left( R_{2,k}^2 \right) + \frac{1}{N_c}\sum_{k=1}^{N_c}\left( R_{3,k}^2 \right),
\end{equation}
where $N_c$ is the number of collocation points.
For this example, 20,000 collocation points
have been used.
The RMSprop optimizer is run for 20,000 iterations. For L-BFGS, maximum allowed 
iteration is set to be 50,000.
For generating benchmark solutions, 
the \texttt{MATLAB}-inbuilt ODE45 function is coupled with \texttt{FERUM}.
The proposed PI-DNN, on the other hand, needs no
such simulator.

The results obtained using the proposed approach
and other state-of-the art methods from the literature are shown in \autoref{tab:eg3_res}.
The threshold value $e_{3p,0}$ is set to be 0.54 and the reliability is evaluated at $t=5.0$.
It is observed that the results obtained using 
the proposed approach matches exactly with the 
MCS results.
Among the other methods, DS yields the most accurate results followed by SORM.
Because of the nonlinear nature of the limit-state function, results obtained using 
FORM are found to be little bit erroneous.
Similar to the previous examples, the number 
of simulations required using different methods are also presented.
FORM, SORM, IS, DS and SS are found to take
112, 139, 1000, 6017 and 1000 runs of the actual solver.
The proposed approach, as already mentioned, needs no simulations.

\begin{table}[ht!]
    \caption{Reliability analysis results for cell-signalling cascade problem.}
    \label{tab:eg3_res}
    \centering
    \begin{tabular}{lcccc}
    \hline
    \textbf{MCS} & $P_f$ & $\beta$ & $N_s$ & $\epsilon = \frac{\left| \beta_e - \beta \right|}{\beta_e}$ \\ \hline
    MCS & 0.0459 & 1.6860 & 10,000 & -- \\ \hline \hline 
    FORM & 0.0750 & 1.4390 & 112 & 14.65\%\\
    SORM & 0.0045 & 1.69 & 139 & 0.23\%   \\
    IS & 0.0467 & 1.6777 & 1000 & 0.49\%   \\
    DS & 0.0455 & 1.6895 & 6017 & 0.21\% \\
    SS & 0.0414 & 1.7347  & 1000 & 2.89\%  \\
    PI-DNN & 0.0459 & 1.6860 & 0 & 0.0\% \\ \hline
    \end{tabular}
\end{table}

\begin{figure}[ht!]
    \centering
    \includegraphics[width = 0.6\textwidth]{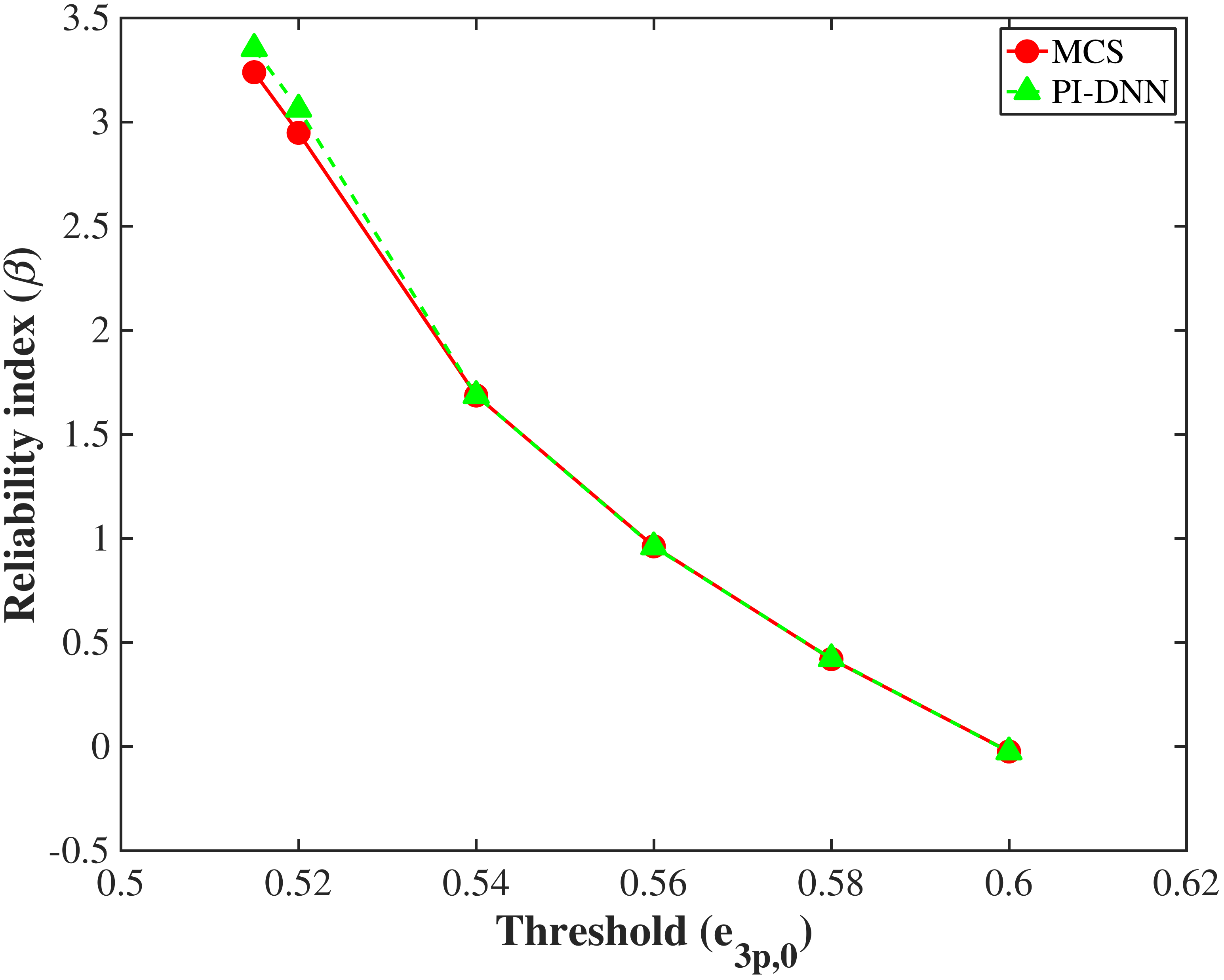}
    \caption{Reliability index corresponding to different threshold $e_{3p,0}$ obtained using MCS and PI-DNN.}
    \label{fig:eg3_thres_vary}
\end{figure}

Similar to \autoref{subsec:burger}, the performance of the proposed PI-DNN in predicting the reliability index corresponding to different thresholds is examined.  
The results obtained are shown in
\autoref{fig:eg3_thres_vary}.
Results corresponding to MCS are also shown.
For all the thresholds, results obtained 
using PI-DNN are found to be extremely close to the MCS predicted results.
This illustrates that the PI-DNN has accurately tracked the response of the stochastic system.

\begin{figure}[ht!]
    \centering
    \subfigure[]{
    \includegraphics[width = 0.48\textwidth]{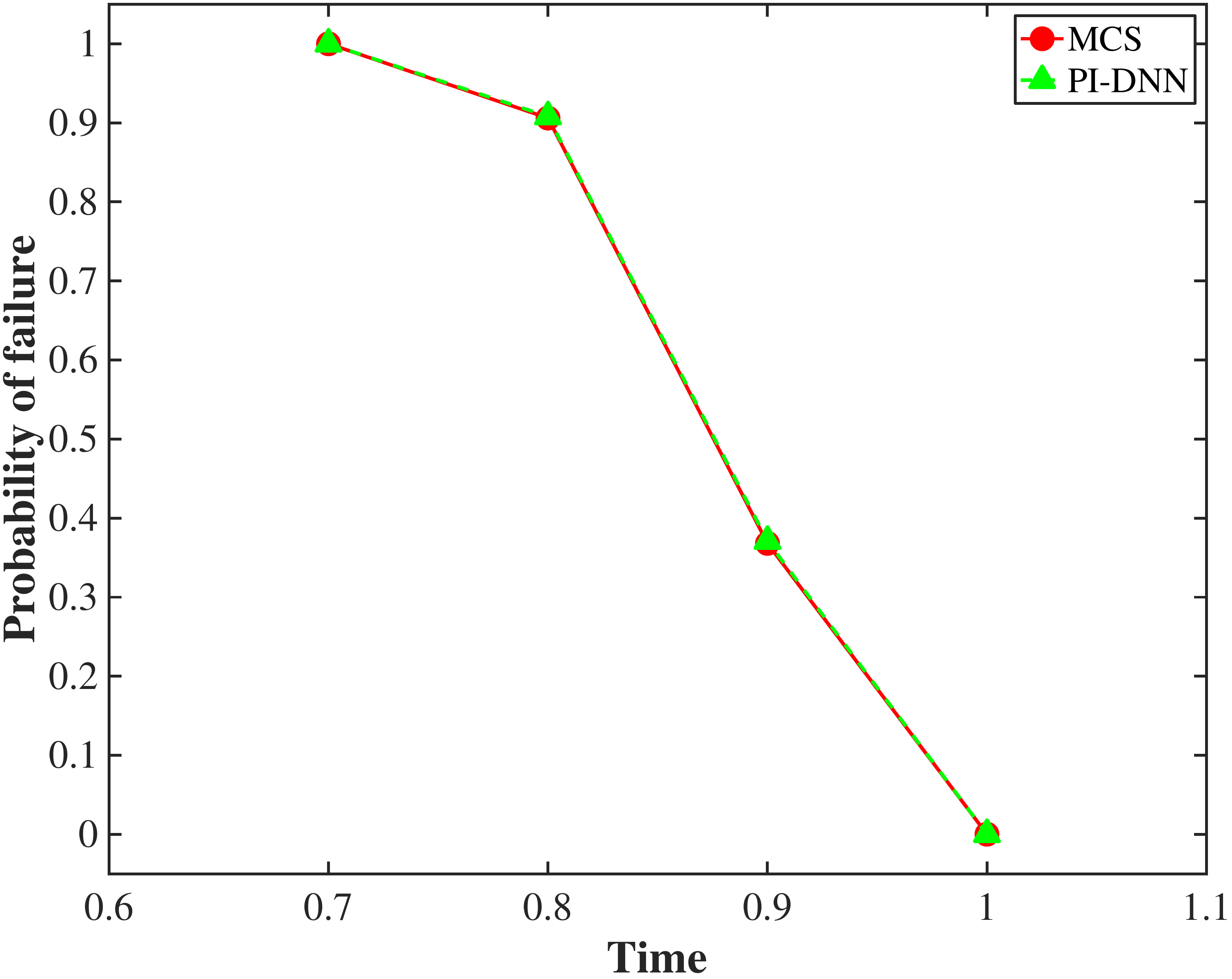}}
    \subfigure[]{
    \includegraphics[width = 0.48\textwidth]{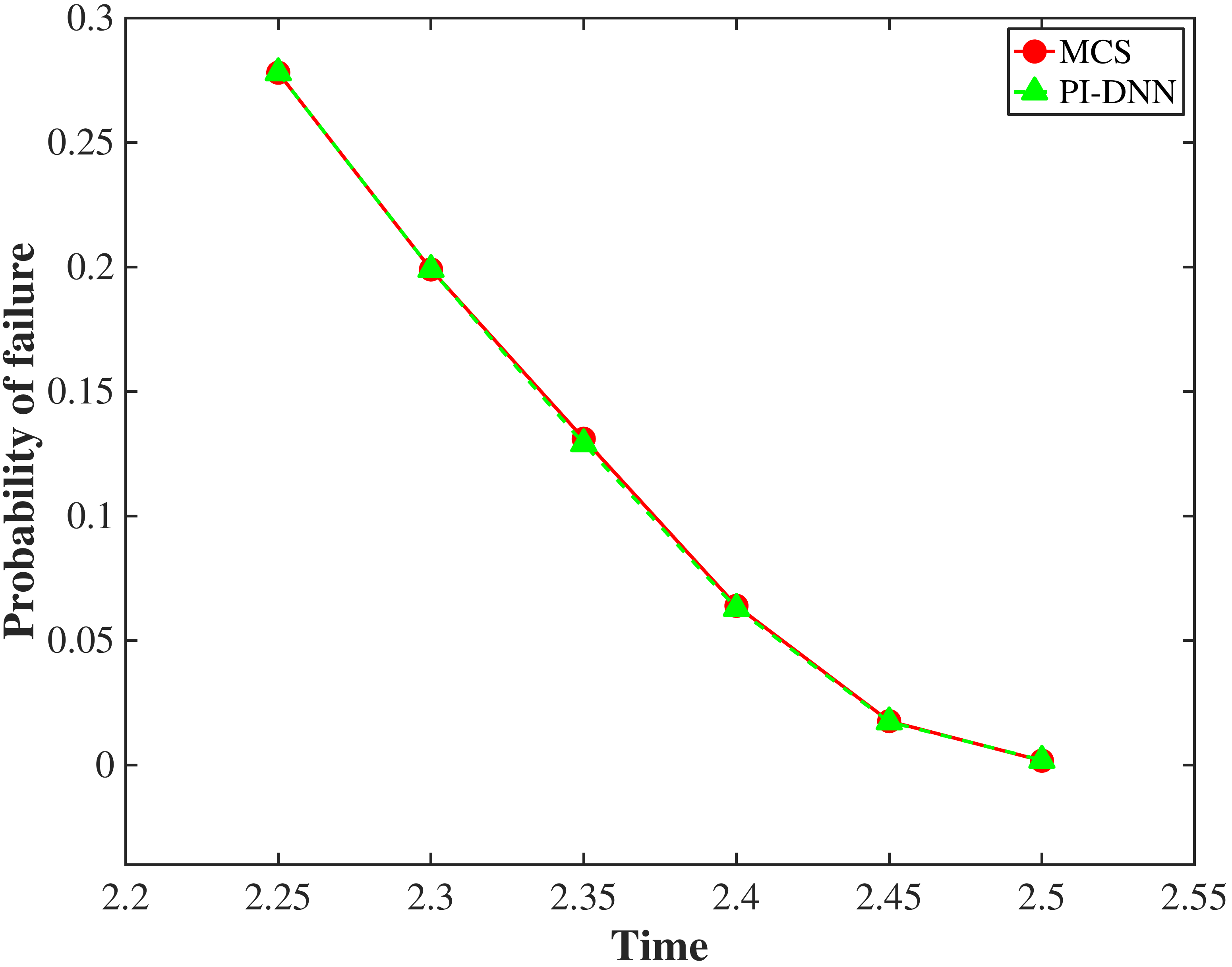}}
    \caption{Probability of failure obtained using the proposed PI-DNN and MCS at different time-instances. $z_0$ for (a) is 0.158 and that for (b) is 0.33. The proposed approach is also able to capture small failure probabilities in the order of $10^-4$.}
    \label{fig:eg3_time}
\end{figure}

One of the interesting features of the proposed
PI-DNN is its ability to predict response
at any point in time and space; 
this is really useful when it comes to 
solving time-dependent reliability analysis problems.
\autoref{fig:eg3_time} shows the probability of failure predicted using the proposed PI-DNN
at different time-steps.
It is observed that the proposed approach
yields reasonably accurate results at all the time-step. Note that all the reliability index
estimates are obtained by using a single 
PI-DNN model.
This indicates the utility of the proposed PI-DNN in solving time-dependent reliability 
analysis problems.

Finally, the influence of network architecture 
and number of collocation points 
on the performance of the proposed approach
is investigated.
\autoref{tab:eg3_cs1} shows the predicted results corresponding to different collocation points.
With increase in number of collocation points,
the proposed PI-DNN is found to stabilize.
\begin{table}[ht!]
    \caption{Reliability analysis results for cell-signaling cascade problem with change in number of collocation points.}
    \label{tab:eg3_cs1}
    \centering
    \begin{tabular}{l||ccccc}
    \hline
        $N_c$ & 5000 & 10000 & 15000 & 20000 & 30000 \\ \hline
         $P_f$ & 0.0525 & 0.0433 & 0.0435 & 0.0459 & 0.0459  \\
         $\beta$ & 1.6211 & 1.7136 & 1.7114 & 1.6860 & 1.6860 \\\hline
    \end{tabular}
\end{table}
\autoref{tab:eg3_cs2} shows the results with
change in the number of hidden layers and number of
neurons.
The results obtained are found to be more or less stable with change in the number of hidden layers and neurons 

\begin{table}[ht!]
    \caption{Variation in reliability analysis results due to change in number of neurons and number of layers (hidden). The numbers in the bracket indicates reliability index}
    \label{tab:eg3_cs2}
    \centering
    \begin{tabular}{l||ccc}
    \hline
    \backslashbox{Layers}{Neurons} & 40 & 50 & 60 \\ \hline \hline
    2   & 0.0469 (1.6576) &	0.0475 (1.6696) & 0.0477 (1.6676) \\ 
    3   & 0.0487 (1.6576) &	0.0458 (1.6870) & 0.0455 (1.6901) \\
    4   & 0.0469 (1.7018) &	0.0459 (1.6860) & 0.0458 (1.6870)  \\\hline 
    \end{tabular}
\end{table}

\section{Conclusions}\label{sec:conclusions}
In this paper, a new class of reliability analysis method that needs {\it no simulation data} is proposed.
The method proposed is based on recent developments in the field of deep learning and
utilizes tools such as automatic differentiation and deep neural networks.
Within the proposed framework, the unknown response is represented by using a deep neural
network. For obtaining the unknown parameters 
of the deep neural network, a physics-informed
loss function is used.
With this loss function, no training data is required and 
the neural network parameters are directly
computed from the governing ODE/PDE of the system.
There are three primary advantages of the proposed approach.
First, the proposed approach needs no simulation data; this is expected to significantly reduce the
computational cost associated with solving
a reliability analysis problems.
Second, since the network parameters are trained by using a physics-informed loss function, physical laws such as invariances and conservation laws will be respected by the 
neural network solution.
Last but not the least, the proposed approach
provides prediction at all spatial and temporal locations and hence, is naturally suited for solving time-dependent reliability analysis problems.

The proposed approach is used for solving three benchmark reliability analysis problems selected
from the literature. 
For all the examples, the proposed approach is found to yield highly accurate results.
Comparison carried out with respect to other state-of-the-art reliability analysis methods indicates the suitability of the proposed approach for solving reliability analysis problems.
Case studies are carried out to investigate convergence of the proposed approach with 
respect to number of collocation points and 
network architecture. 
The results obtained indicate that the stability and robustness of the proposed approach.

It is to be noted that the approach presented in this paper can further be enhanced in number of ways.
For example, replacing the fully connected deep
neural network with a convolutional type neural
network will possibly enable us to solve really high dimensional reliability analysis problems.
Similarly, there is a huge scope to develop adaptive version of this algorithm that will select collocation points in an adaptive manner.
The neural network architecture can also be selected by using an adaptive framework. Some of this possibilities will be explored in future studies.
\section*{Acknowledgements}
The author thanks Govinda Anantha Padmanabhaa for his help with generating the benchmark results for problem 2. 
The author also thanks Soumya Chakraborty for proof reading this article.
The \texttt{TensorFlow} codes were run on google colab service.


\end{document}